\documentclass{article}

\usepackage[preprint]{neurips_2022}

\usepackage[utf8]{inputenc} 
\usepackage[T1]{fontenc}    
\usepackage{hyperref}       
\usepackage{url}            
\usepackage{booktabs}       
\usepackage{amsfonts}       
\usepackage{nicefrac}       
\usepackage{microtype}      
\usepackage[dvipsnames]{xcolor}         

\usepackage{amssymb}
\usepackage{amsthm}
\usepackage{amsmath}
\usepackage{graphicx}
\usepackage{wrapfig}
\usepackage[shortlabels,inline]{enumitem}
\usepackage{cleveref}
\setlist{nosep}

\title{Exploring the Representation Manifolds of Stable Diffusion Through the Lens of Intrinsic Dimension}

\author{%
  Henry Kvinge\thanks{H.K has a joint appointment in the University of Washington Department of Mathematics.}, Davis Brown, Charles Godfrey \\
  Pacific Northwest National Laboratory\\
  \texttt{\{first\}.\{last\}@pnnl.gov} \\
}

\begin{document}

\maketitle

\begin{abstract}
Prompting has become an important mechanism by which users can more effectively interact with many flavors of foundation model. Indeed, the last several years have shown that well-honed prompts can sometimes unlock emergent capabilities within such models. While there has been a substantial amount of empirical exploration of prompting within the community, relatively few works have studied prompting at a mathematical level. In this work we aim to take a first step towards understanding basic geometric properties induced by prompts in Stable Diffusion, focusing on the intrinsic dimension of internal representations within the model. We find that choice of prompt has a substantial impact on the intrinsic dimension of representations at both layers of the model which we explored, but that the nature of this impact depends on the layer being considered. For example, in certain bottleneck layers of the model, intrinsic dimension of representations is correlated with prompt perplexity (measured using a surrogate model), while this correlation is not apparent in the latent layers. Our evidence suggests that intrinsic dimension could be a useful tool for future studies of the impact of different prompts on text-to-image models.
\end{abstract}

\section{Introduction}

The current generation of generative text-to-image models (e.g., DALLE-2 \citep{ramesh2022hierarchical}, Stable Diffusion \citep{Rombach_2022_CVPR}, Parti \citep{yuscaling}) are remarkable in their responsiveness to complex textual input. While this is easy to observe in practice by simply interacting with the models, the number of works trying to quantify the way that text prompts shape and shift model output have so far been limited. Yet, as prompting becomes an important way by which humans interact with foundation models, there is a need for quantitative methods of measuring and understanding its impact on the behavior and function of a model. In this work we take a first step in this direction by studying the hidden representations within Stable Diffusion induced by varying prompts through the lens of essential geometric features.

We focus on understanding how prompts affect the intrinsic dimension of distributions of hidden activations from several points within the model. The well-known Manifold Hypothesis posits that many types of high-dimensional signals actually live on low-dimensional manifolds. The most fundamental property of a manifold is arguably its dimension and thus there are a range of methods of estimating the dimension of the manifold underlying a finite number of points  (usually termed the {\emph{intrinsic dimension}} of the data). A range of works have studied deep learning models and their behavior using intrinsic dimension, from model generalization \citep{pope2021the} to adversarial vulnerability \citep{amsaleg2017vulnerability}. In this work we run a range of experiments to better understand how properties of a prompt translate into geometric features of the distribution of internal representations, eventually resulting in dazzling and varied visual output. Our goal is to ultimately be able to shed light on the prompting process, leading us to understand how to get better performance out of our models.

Among our findings are the following. (i) Prompts have a substantial effect on the distribution of internal representations. Two similarly generic prompts can result in distributions of representations that vary by up to 10 dimensions. (ii) In the case of Stable Diffusion, where images are generated from an iterative diffusion process, we in fact get a \emph{sequence} of output distributions and we find that the dynamics of this sequence depends on what part of the model representations are being extracted from. (iii) For representations extracted from the middle of the model (specifically in the UNet bottleneck), intrinsic dimension correlates with prompt perplexity as measured by a surrogate language model. In words, when given more obscure prompts, text-to-image models generate distributions with support on a higher-dimensional manifold. While we restrict attention to Stable Diffusion here, we hope that this work will be a precursor for more comprehensive studies of other models that utilize prompts (such as large language models). 

\begin{figure}[h]
\begin{center}
\includegraphics[width=0.48\linewidth]{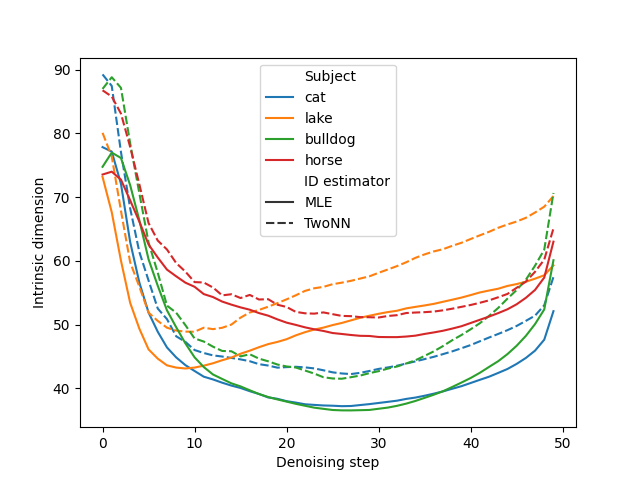}
\includegraphics[width=0.48\linewidth]{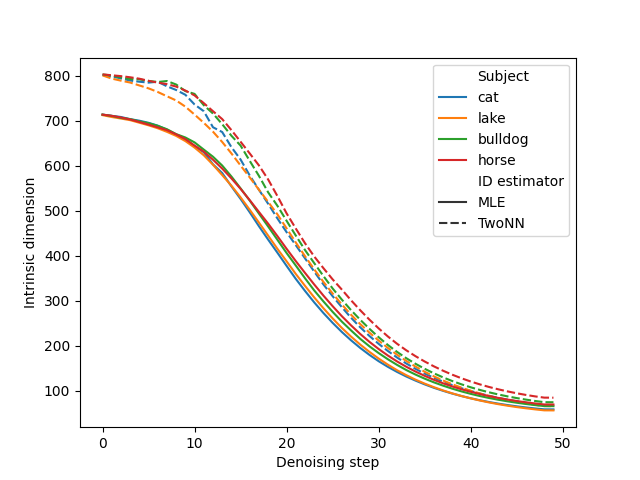}
\end{center}
\caption{Estimations of the intrinsic dimensions of the bottleneck representation (\textbf{left}) and latent representation (\textbf{right}) as a function of denoising step. We use MLE (solid line) and TwoNN (dashed line) to estimate intrinsic dimension. The prompt `{\color{blue}{cat}}' is short for `{\color{blue}{a cute cat}}', the prompt `{\color{orange}{lake}}' is short for `{\color{orange}{a photograph of a beautiful lake}}', the prompt `{\color{ForestGreen}{bulldog}}' is short for `{\color{ForestGreen}{a painting of a french bulldog in the rain}}', and the prompt `{\color{red}{horse}}' is short for `{\color{red}{a photograph of an astronaut riding a horse}}'.} \label{fig-vanilla-intrinsic-dimension}
\end{figure}

\section{A manifold framework for prompts}

We start by establishing a geometric framework for the output of text-to-image diffusion models. Let $\mathcal{D}$ be a noise distribution in $\mathbb{R}^n$, and let $\mathcal{P}$ be a distribution on the set of all discrete text strings $P$. Stable Diffusion, which we denote by $S: \mathbb{R}^n \times P \rightarrow \mathbb{R}^{3 \times h \times w}$, is a continuous map that takes two inputs: noise $n$ drawn from $\mathcal{D}$ in $\mathbb{R}^n$ and a text string drawn from $p \in P$ and produces a size $h \times w$ RGB image. We assume that $S$ is the composition $S = S_\ell \circ S_{\ell-1} \circ \dots \circ S_2 \circ S_1$, where each $S_i: \mathbb{R}^{n_i} \rightarrow \mathbb{R}^{n_{i+1}}$ for $1 \leq i \leq \ell$ is a layer of the model. We denote by $S_{\leq i} = S_{i} \circ \dots \circ S_{1}$ the composition of the first $i$ layers of the model and by $S_{>i} = S_{\ell} \circ \dots \circ S_{i+1}$ the last $\ell - i$ layers of the model.

Let $\mathcal{X}_i$ be the pushforward of the product of distributions $\mathcal{D} \times \mathcal{P}$ under the map $S_{\leq i}$, $S_{\leq i}(\mathcal{D},\mathcal{P})$. In the spirit of the Manifold Hypothesis, we assume that the support of distribution $S_{\leq i}(\mathcal{P},\mathcal{D})$ is a manifold in $\mathbb{R}^{n_{i+1}}$ which we call the {\emph{representation manifold at layer $i$}}, $X_i$ (that is, we get a manifold embedded in $\mathbb{R}^{n_{i+1}}$ if we apply $S_{\leq i}$ to all possible noise $n$ and prompts $p$). For fixed $p \in P$, we can use a similar construction to get a submanifold $X_i^p$ of $X_i$ corresponding to $p$ (see Figure \ref{fig-cartoon-visualization} for a cartoon of these constructions). 

Note that, being a diffusion model, $S$ contains an inner loop corresponding to denoising. As such, if layer $i$ is iterated over in denoising, we actually get a sequence of manifolds $\{X_i^{p,1},X_i^{p,2},....\}$, where $X_i^{p,t}$ is the manifold obtained at layer $i$, for prompt $p$, at the $t$-th step of denoising. In this paper we study the intrinsic dimension of $X_i^{p,t}$ for varying $i$, $p$, and $t$. By abuse of notation we write $X_i^{p}$ to denote $X_i^{p,T}$, where $T$ is the final denoising step.

In our experiments we record hidden activations from two parts of the model, the latent space (prior to applying the decoder component of the VAE) and the output of the layer \verb|unet.down_blocks[3].resnets[1].nonlinearity| which is near the bottleneck of the UNet. We refer to this latter representation as the {\emph{bottleneck representation}}, though the output of other layers could also be called this.

\begin{figure}[h]
\begin{center}
\includegraphics[width=0.75\linewidth]{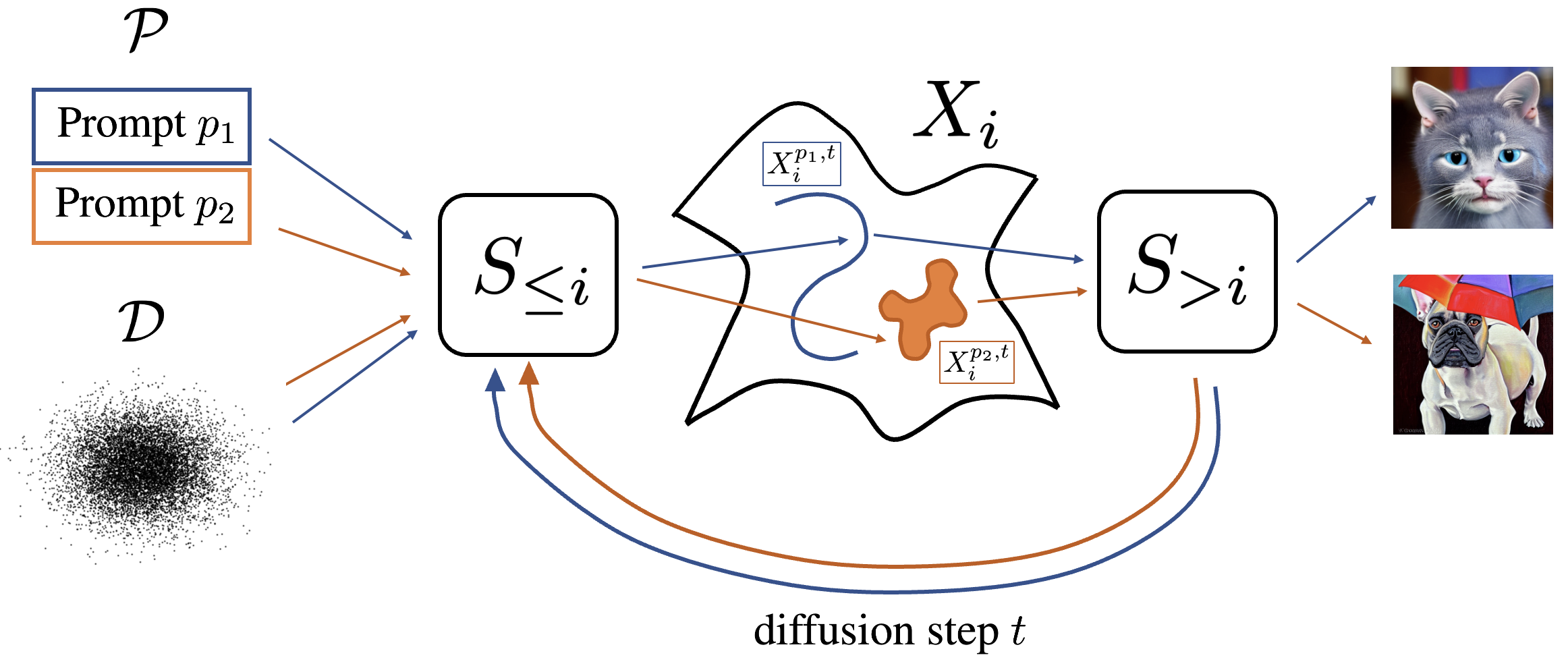}
\end{center}
\caption{A illustration of the framework with which we work. Recall that $S_{\leq i}$ are the first $i$ layers of Stable Diffusion and $S_{> i}$ are the remaining layers. For simplicity, we mostly obscure the different components of the model (e.g., the text encoder and VAE decoder).} \label{fig-cartoon-visualization}
\end{figure}

\section{Experiments}
\label{sec:exper}

We work with Stable Diffusion v1-4 found at Hugging Face\footnote{https://huggingface.co/CompVis/stable-diffusion-v1-4} and describe hyperparameters and other experimental details in Section \ref{appendix-experimental-details} in the Appendix. We use two popular intrinsic dimension estimators: maximum likelihood estimator (MLE) \citep{mle} and Two-NN \citep{facco2017estimating} which we describe in Section \ref{appendix-ID} of the Appendix. In the remainder of this paper, we ask a range of questions about how differences and properties of prompts impact the intrinsic dimension of both the manifold of latent representations and the manifold of bottleneck representations.\\

\textbf{Does the prompt impact the dimension of representation manifolds?:} As a baseline, in our first series of experiments we measure the extent to which intrinsic dimension varies between four generic prompts. To do this we generated 5,000 images with for each of the prompts: `a cute blue cat', `a photograph of a beautiful lake', `a painting of a French bulldog in the rain', and `a photograph of an astronaut riding a horse.' We recorded both the bottleneck and latent representations of all 5,000 instances of each class at all 50 denoising steps and then calculated the intrinsic dimension of these 50 $\times$ 4 $=$ 200 point clouds using both MLE and TwoNN. The results, plotted as a function of denoising step are shown in Figure \ref{fig-vanilla-intrinsic-dimension} with the intrinsic dimensions of the final denoising step (denoising step 50) shown in Figure \ref{fig-vanilla-intrinsic-dimension-final} in the Appendix. 

When viewed as a function of denoising step, we find that the latent representation intrinsic dimension decreases monotonically while the intrinsic dimension of the bottleneck representations initially decreases and then increases, forming a {\emph{U}}-shape. In Section \ref{appendix-why-does-intrinsic-dimension-increase} of the Appendix, we provide evidence that this increase in dimension for bottleneck representation manifolds toward the end of denoising can be traced to the time step embedding vector, which adds increasingly high-frequency signal into the representations. Indeed, even if we freeze the input to the UNet at 40 steps (so that the input for steps 41 through 50 is constant) the intrinsic dimension still increases, indicating that this is due to some factor independent of the denoising process itself. 

Beyond their denoising dynamics, the $4$ different prompts resulted in representation manifolds with significantly different intrinsic dimension. For example, the latent representation (right plot in Figure \ref{fig-vanilla-intrinsic-dimension-final}) ranged from $~55$ to $~70$ (MLE) to $~65$ to $~85$ (TwoNN). For latent representations, the relative ordering of the different prompts in terms of intrinsic dimension is fairly consistent between estimators (MLE vs TwoNN), whereas this is less true for bottleneck representations where different methods give substantially less consistent results. This may point to more geometric and topological complexity in bottleneck representations, but further experiments would be needed to provide evidence of this conjecture. 

\textbf{Takeaway:} Intrinsic dimension in latent representation manifolds decreases over the course of denoising whereas the curve for intrinsic dimension of the bottleneck representations forms a U-shaped curve (likely due to the time step embedding vector). Different prompts tend to produce representation manifolds with substantially different intrinsic dimensions.

\begin{wrapfigure}{r}{0.5\textwidth}
\begin{center}
\includegraphics[width=0.9\linewidth]{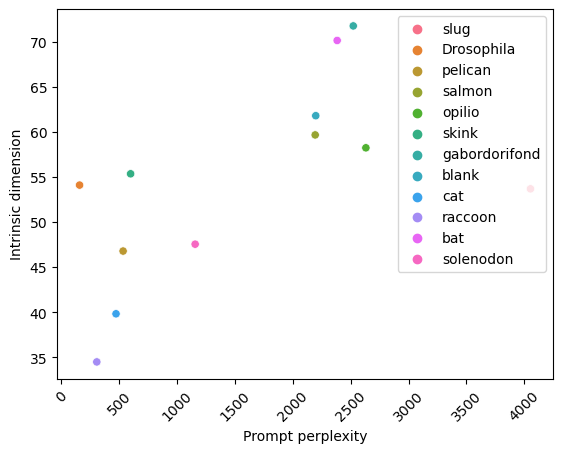}
\end{center}
\caption{Plots showing the perplexity of the prompt `A cute --' (where '--' is replaced by each of the words in the legend, e.g., `slug') vs. the intrinsic dimension (measured in terms of MLE) of the 5,000 bottleneck representations produced by this prompt.} \label{fig-perplexity}
\end{wrapfigure}

\textbf{What properties of a prompt impact the dimension of representation manifolds?} We consider the phrase `A cute --' where we replace `--' with each of the following: `slug', `Drosophila', `pelican', `salmon', `opilio', `skink',   `gabordorifond', `raccoon', `bat', `solenodon'\footnote{Note that Drosophila is the shortened name of {\emph{Drosophila melanogaster}}, a species of fruit fly that has been the model animal in many areas of biology, an opilio is a species of snow crab, a skink is a lizard belonging to the family Scincidae, `gabordorifond' is a made up word, and a solenodon is a venomous rodent.}. These animal were chosen so that some were likely to be associated with the phrase `A cute --' and some were not (e.g., slug). We measured the perplexity of each prompt using a surrogate language model --- the underlying hypothesis here being that less likely prompts occur less frequently (if at all) in the training data of the surrogate language model, and will thus incur higher perplexity. Comparing the perplexity scores with the actual frequencies of prompt tokens in LAION-5B \citep{schuhmann2022laion5b} is an interesting avenue for future work. Figure \ref{fig-perplexity} displays scatter plots of the intrinsic dimension and perplexity of these prompts. 

Strikingly for the bottleneck representation manifold $X_i^p$ there is a noticeable correlation between the perplexity of $p$ and intrinsic dimension of $X_i^p$ (Figure \ref{fig-perplexity}). This is consistent with an intuitive hypothesis that more unlikely prompts are out-of-distribution for the Stable Diffusion model, resulting in a noisier distribution of representations and in turn a higher intrinsic dimension estimate of the underlying manifold. This observation can be roughly related to the functional form of the neural scaling law found in \cite{bahriExplainingNeuralScaling2021a}; in the ``resolution limited'' regime where the number of parameters is effectively infinite, they find that the loss $L = \mathcal{O}(D^{-\frac{1}{d}})$ where \(D\) is the number of datapoints and \(d \) the intrinsic dimension of the data manifold. Hence with \(D \) fixed, the intrinsic dimension $d$ is monotonically related to the loss $L$ (i.e., roughly the perplexity) --- see \cref{sec:envelope} for further discussion. However this hypothesis must be treated with caution, as there seems to be almost no correlation between perplexity and intrinsic dimension in the latent representations (Figure \ref{fig-perplexity-latent} in the Appendix). Note that Figure \ref{fig-vanilla-intrinsic-dimension} also shows the bottleneck and latent representations can behave quite differently from the perspective of intrinsic dimension. 

\textbf{Takeaway:} The intrinsic dimension of bottleneck representation manifolds correlates with the perplexity of a prompt (a proxy for the occurrence of the prompt in the training data). 


\section{Conclusion}

In this paper we show that the choice of input prompt to Stable Diffusion has a substantial effect on the intrinsic dimension of the resulting internal representations within the model. We find evidence that, at least at bottleneck layers of the UNet component of the model, prompt perplexity is correlated with the intrinsic dimension of internal representations. We hope that this is a small first step to establishing a better understanding of the geometry of learning in large text-to-image models. 

\begin{ack}
This research was supported by the Mathematics for Artificial Reasoning in Science (MARS) initiative at Pacific Northwest National Laboratory.
It was conducted under the Laboratory Directed Research and Development (LDRD) Program at at Pacific Northwest National Laboratory (PNNL), a multiprogram
National Laboratory operated by Battelle Memorial Institute for the U.S. Department of Energy under Contract
DE-AC05-76RL01830.
\end{ack}

\bibliography{refs}

\begin{thebibliography}{18}
\providecommand{\natexlab}[1]{#1}
\providecommand{\url}[1]{\texttt{#1}}
\expandafter\ifx\csname urlstyle\endcsname\relax
  \providecommand{\doi}[1]{doi: #1}\else
  \providecommand{\doi}{doi: \begingroup \urlstyle{rm}\Url}\fi

\bibitem[Aghajanyan et~al.(2023)Aghajanyan, Yu, Conneau, Hsu, Hambardzumyan,
  Zhang, Roller, Goyal, Levy, and
  Zettlemoyer]{aghajanyanScalingLawsGenerative2023}
Armen Aghajanyan, Lili Yu, Alexis Conneau, Wei-Ning Hsu, Karen Hambardzumyan,
  Susan Zhang, Stephen Roller, Naman Goyal, Omer Levy, and Luke Zettlemoyer.
\newblock Scaling {{Laws}} for {{Generative Mixed-Modal Language Models}},
  January 2023.

\bibitem[Amsaleg et~al.(2017)Amsaleg, Bailey, Barbe, Erfani, Houle, Nguyen, and
  Radovanovi{\'c}]{amsaleg2017vulnerability}
Laurent Amsaleg, James Bailey, Dominique Barbe, Sarah Erfani, Michael~E Houle,
  Vinh Nguyen, and Milo{\v{s}} Radovanovi{\'c}.
\newblock The vulnerability of learning to adversarial perturbation increases
  with intrinsic dimensionality.
\newblock In \emph{2017 ieee workshop on information forensics and security
  (wifs)}, pp.\  1--6. IEEE, 2017.

\bibitem[Ansuini et~al.(2019)Ansuini, Laio, Macke, and
  Zoccolan]{ansuini2019intrinsic}
Alessio Ansuini, Alessandro Laio, Jakob~H Macke, and Davide Zoccolan.
\newblock Intrinsic dimension of data representations in deep neural networks.
\newblock \emph{Advances in Neural Information Processing Systems}, 32, 2019.

\bibitem[Bahri et~al.(2021)Bahri, Dyer, Kaplan, Lee, and
  Sharma]{bahriExplainingNeuralScaling2021a}
Yasaman Bahri, Ethan Dyer, Jared Kaplan, Jaehoon Lee, and Utkarsh Sharma.
\newblock Explaining {{Neural Scaling Laws}}, February 2021.

\bibitem[Caballero et~al.(2023)Caballero, Gupta, Rish, and
  Krueger]{caballeroBrokenNeuralScaling2023}
Ethan Caballero, Kshitij Gupta, Irina Rish, and David Krueger.
\newblock Broken {{Neural Scaling Laws}}, January 2023.

\bibitem[Cherti et~al.(2022)Cherti, Beaumont, Wightman, Wortsman, Ilharco,
  Gordon, Schuhmann, Schmidt, and Jitsev]{chertiReproducibleScalingLaws2022}
Mehdi Cherti, Romain Beaumont, Ross Wightman, Mitchell Wortsman, Gabriel
  Ilharco, Cade Gordon, Christoph Schuhmann, Ludwig Schmidt, and Jenia Jitsev.
\newblock Reproducible scaling laws for contrastive language-image learning,
  December 2022.

\bibitem[Choi et~al.(2022)Choi, Hwang, Cho, and Kang]{choi2022analyzing}
Jaewoong Choi, Geonho Hwang, Hyunsoo Cho, and Myungjoo Kang.
\newblock Analyzing the latent space of gan through local dimension estimation.
\newblock \emph{arXiv preprint arXiv:2205.13182}, 2022.

\bibitem[Facco et~al.(2017)Facco, d’Errico, Rodriguez, and
  Laio]{facco2017estimating}
Elena Facco, Maria d’Errico, Alex Rodriguez, and Alessandro Laio.
\newblock Estimating the intrinsic dimension of datasets by a minimal
  neighborhood information.
\newblock \emph{Scientific reports}, 7\penalty0 (1):\penalty0 1--8, 2017.

\bibitem[Gong et~al.(2019)Gong, Boddeti, and
  Jain]{gongIntrinsicDimensionalityImage2019}
Sixue Gong, Vishnu~Naresh Boddeti, and Anil~K. Jain.
\newblock On the {{Intrinsic Dimensionality}} of {{Image Representations}},
  April 2019.

\bibitem[Horvat \& Pfister(2022)Horvat and
  Pfister]{horvatIntrinsicDimensionalityEstimation2022}
Christian Horvat and Jean-Pascal Pfister.
\newblock Intrinsic dimensionality estimation using {{Normalizing Flows}}.
\newblock In \emph{Advances in {{Neural Information Processing Systems}}},
  October 2022.

\bibitem[Levina \& Bickel(2004)Levina and Bickel]{mle}
Elizaveta Levina and Peter~J. Bickel.
\newblock Maximum likelihood estimation of intrinsic dimension.
\newblock In \emph{Proceedings of the 17th International Conference on Neural
  Information Processing Systems}, NIPS'04, pp.\  777–784, Cambridge, MA,
  USA, 2004. MIT Press.

\bibitem[Pope et~al.(2021)Pope, Zhu, Abdelkader, Goldblum, and
  Goldstein]{pope2021the}
Phil Pope, Chen Zhu, Ahmed Abdelkader, Micah Goldblum, and Tom Goldstein.
\newblock The intrinsic dimension of images and its impact on learning.
\newblock In \emph{International Conference on Learning Representations}, 2021.
\newblock URL \url{https://openreview.net/forum?id=XJk19XzGq2J}.

\bibitem[Ramesh et~al.(2022)Ramesh, Dhariwal, Nichol, Chu, and
  Chen]{ramesh2022hierarchical}
Aditya Ramesh, Prafulla Dhariwal, Alex Nichol, Casey Chu, and Mark Chen.
\newblock Hierarchical text-conditional image generation with clip latents.
\newblock \emph{arXiv preprint arXiv:2204.06125}, 2022.

\bibitem[Rombach et~al.(2022)Rombach, Blattmann, Lorenz, Esser, and
  Ommer]{Rombach_2022_CVPR}
Robin Rombach, Andreas Blattmann, Dominik Lorenz, Patrick Esser, and Bj\"orn
  Ommer.
\newblock High-resolution image synthesis with latent diffusion models.
\newblock In \emph{Proceedings of the IEEE/CVF Conference on Computer Vision
  and Pattern Recognition (CVPR)}, pp.\  10684--10695, June 2022.

\bibitem[Schuhmann et~al.(2022)Schuhmann, Beaumont, Vencu, Gordon, Wightman,
  Cherti, Coombes, Katta, Mullis, Wortsman, Schramowski, Kundurthy, Crowson,
  Schmidt, Kaczmarczyk, and Jitsev]{schuhmann2022laion5b}
Christoph Schuhmann, Romain Beaumont, Richard Vencu, Cade Gordon, Ross
  Wightman, Mehdi Cherti, Theo Coombes, Aarush Katta, Clayton Mullis, Mitchell
  Wortsman, Patrick Schramowski, Srivatsa Kundurthy, Katherine Crowson, Ludwig
  Schmidt, Robert Kaczmarczyk, and Jenia Jitsev.
\newblock Laion-5b: An open large-scale dataset for training next generation
  image-text models.
\newblock \emph{NEURIPS}, 2022.

\bibitem[Sharma \& Kaplan(2020)Sharma and Kaplan]{sharmaNeuralScalingLaw2020}
Utkarsh Sharma and Jared Kaplan.
\newblock A {{Neural Scaling Law}} from the {{Dimension}} of the {{Data
  Manifold}}, April 2020.

\bibitem[Stanczuk et~al.(2023)Stanczuk, Batzolis, and
  Sch{\"o}nlieb]{stanczukYourDiffusionModel2023}
Jan Stanczuk, Georgios Batzolis, and Carola-Bibiane Sch{\"o}nlieb.
\newblock Your diffusion model secretly knows the dimension of the data
  manifold, January 2023.

\bibitem[Yu et~al.()Yu, Xu, Koh, Luong, Baid, Wang, Vasudevan, Ku, Yang, Ayan,
  et~al.]{yuscaling}
Jiahui Yu, Yuanzhong Xu, Jing~Yu Koh, Thang Luong, Gunjan Baid, Zirui Wang,
  Vijay Vasudevan, Alexander Ku, Yinfei Yang, Burcu~Karagol Ayan, et~al.
\newblock Scaling autoregressive models for content-rich text-to-image
  generation.
\newblock \emph{Transactions on Machine Learning Research}.

\end{thebibliography}
\bibliographystyle{iclr2023_conference}

\appendix

\section{Related work}

Stable Diffusion \citep{Rombach_2022_CVPR} generates images from textual prompts using a diffusion model trained to map random noise to natural images, where the denoising process is \emph{conditioned} on the text prompts. For further details we refer to the original paper. Other notable text-to-image generative models use different architectures, training and inferrence, for example DALL-E \citep{ramesh2022hierarchical} trains a variational autoencoder to map image patches to the same token space as prompt text, and then trains an autoregressive model on sequences of captions followed by images. Adapting the experiments in this paper to the quite different framework of DALL-E is an interesting direction for future work.

Arguably the most essential property of a manifold is its dimension. Many algorithms have been introduced to estimate dimension from point samples; in this work we use those of \cite{mle} and \cite{facco2017estimating}. A number of works have attempted to use generative models to estimate intrinsic dimension of data \citep{horvatIntrinsicDimensionalityEstimation2022,stanczukYourDiffusionModel2023} --- these are more-or-less unrelated to our experiments, where we take a fixed diffusion model and measure intrinsic dimension of certain distributions in its feature spaces with more classical dimension estimators.

\cite{choi2022analyzing} studied the intrinsic dimension of representations in latent spaces of generative adversarial networks (GANs). \cite{pope2021the} computed the intrinsic dimension of image distributions generated by GANs (i.e. GAN outputs, as opposed to latents) as well as the intrinsic dimension of natural image datasets (see also \citep{gongIntrinsicDimensionalityImage2019}). None of these works measures intrinsic dimension on latent spaces of image diffusion models. 

\section{Implementation}

\subsection{Bottleneck representations and latent representations}

In this section we give more details on the layers at which we record hidden activations for our experiments. Latent space representations are the output of the UNet component of the Stable Diffusion model. They are of dimension $4 \times 64 \times 64$ and resemble small versions of the final higher resolution images that are produced by putting them through the final VAE decoder component of the model. The bottleneck representation is the output of layer \verb|unet.down_blocks[3].resnets[1].nonlinearity| in the UNet component of the model. It has shape $2\times 1280 \times 8 \times 8$. We could have also chosen to take the intrinsic dimension of the final output image, which in our set-up had dimensions $3 \times 512 \times 512$, but chose not to because (i) the latent representations seems to already have much of the semantic content (in a lower resolution form) as the latent representation and (ii) the VAE is not conditioned on the prompt (unlike the UNet).

\subsection{Choice of intrinsic dimensionality estimators} \label{appendix-ID}

We use two popular intrinsic dimension estimators: maximum likelihood estimator (MLE) \citep{mle} and Two-NN \citep{facco2017estimating}. Given a dataset $D$, MLE treats the presence of datapoints within a ball $B_r(x)$ of radius $r$, centered at $x \in D$, as a Poisson process. This framework is then used to predict the dimension of the manifold from which $D$ was sampled. Two-NN only uses the two nearest neighbors to estimate the intrinsic dimension. We chose these two methods because they have been successfully applied to data from deep learning models in past works such as \cite{pope2021the} and \cite{ansuini2019intrinsic}. 

\begin{figure}[h]
\begin{center}
\includegraphics[width=0.49\linewidth]{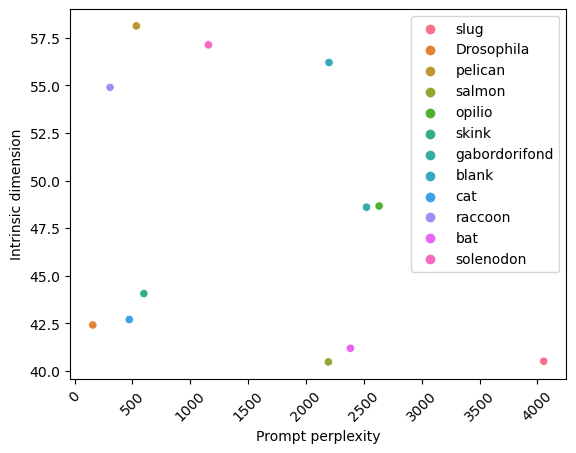}
\end{center}
\caption{Plots showing the perplexity of the prompt `A cute --' (where '--' is replaced by each of the words in the legend, e.g., `slug') vs. the intrinsic dimension (measured in terms of MLE) of the 5,000 latent representations produced by this prompt.} \label{fig-perplexity-latent}
\end{figure}

\begin{figure}[h]
\begin{center}
\includegraphics[width=0.48\linewidth]{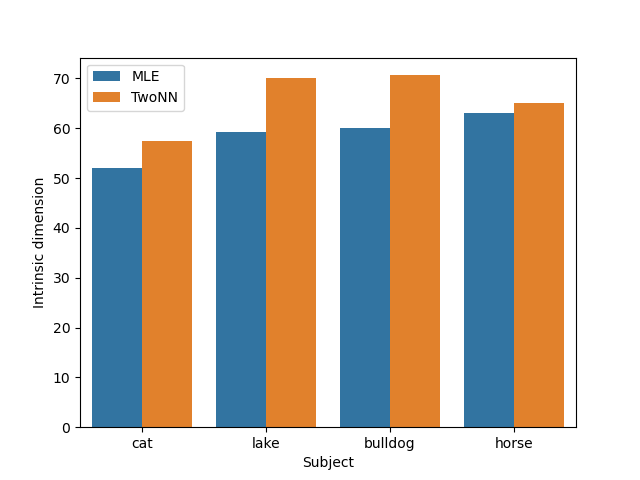}
\includegraphics[width=0.48\linewidth]{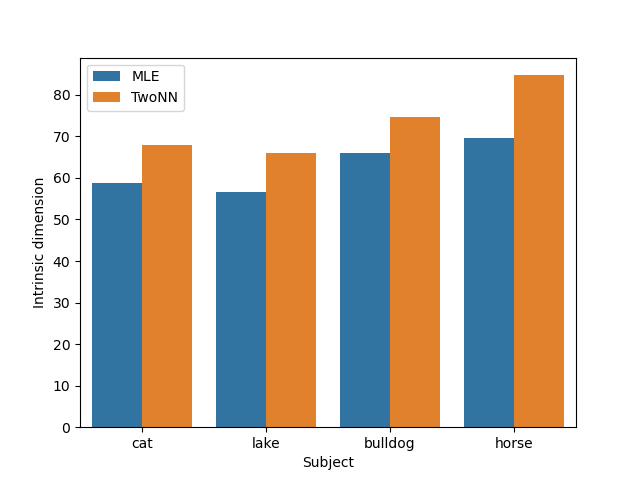}
\end{center}
\caption{Estimations of the intrinsic dimensions of the bottleneck representation manifold (\textbf{left}) and latent representation manifold (\textbf{right}) in the final denoising step. We use {\color{NavyBlue}{MLE}} and {\color{orange}{TwoNN}} to estimate intrinsic dimension. The full prompts can be found in the caption of Figure \ref{fig-vanilla-intrinsic-dimension}.} \label{fig-vanilla-intrinsic-dimension-final}
\end{figure}

\subsection{Experimental details}
\label{appendix-experimental-details}

Unless otherwise stated, in all experiments we use 50 denoising steps with a guidance scale of 7.5. When computing the intrinsic dimension of representations associated with a prompt $p$, we always generate 5,000 images. We found empirically that increasing the number of images only had a small impact on the intrinsic dimension estimates, so we conjecture that the patterns that we observe here persist as the number of images becomes increasingly large.

\begin{figure}[h]
\begin{center}
\includegraphics[width=0.49\linewidth]{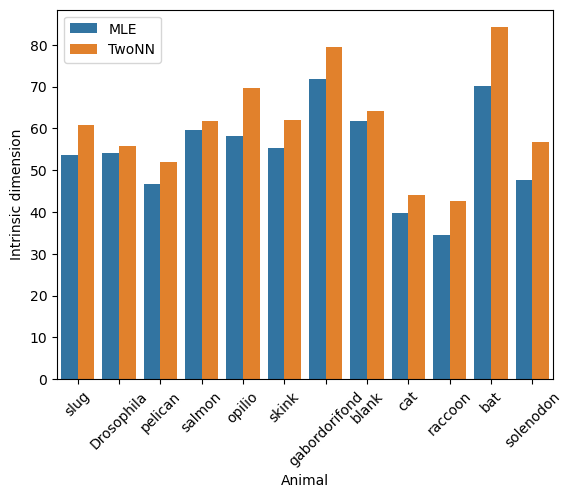}
\includegraphics[width=0.49\linewidth]{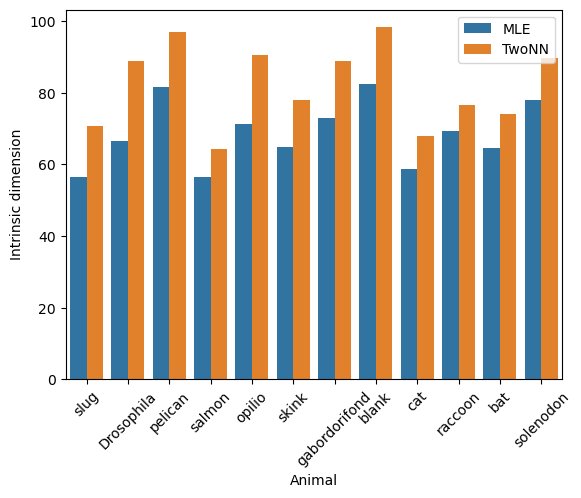}
\end{center}
\caption{The intrinsic dimension of 5,000 bottleneck representations (\textbf{left}) and latent representations (\textbf{right}) with respect to MLE and TwoNN, using prompts of the form `A cute --' where `--' is the name of an animal on the $x$-axis.} \label{fig-cartoon-visualization}
\end{figure}

\section{Why does intrinsic dimension increase at later steps in denoising for bottleneck representations?}
\label{appendix-why-does-intrinsic-dimension-increase}

\begin{figure}
\begin{center}
\includegraphics[width=0.5\linewidth]{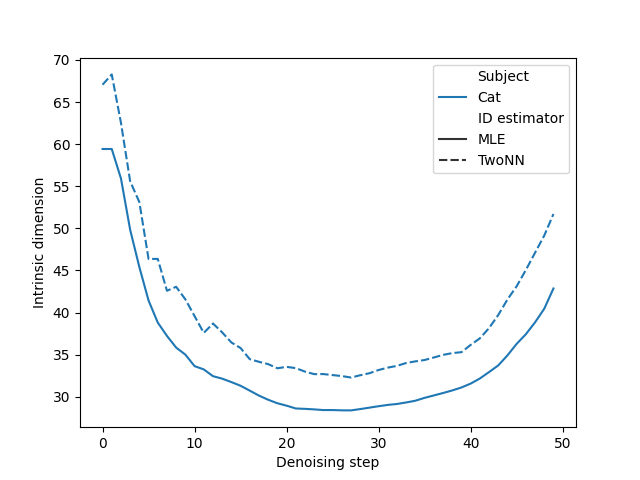}
\end{center}
\caption{The intrinsic dimension of 500 bottleneck representations over 50 denoising steps when input to the UNet is frozen after step 40 (so that the only change is due to the time embedding vector). The intrinsic dimension estimator used was MLE and the prompt was `A cute blue cat'.} \label{fig-time-frozen}
\end{figure}

As can be observed in Figure \ref{fig-vanilla-intrinsic-dimension}, for a fixed prompt $p$, measurements of the intrinsic dimension of the bottleneck representation manifold $X_i^{p,t}$ initially decrease and then increase again. This is in contrast to the analogous statistics of the latent representation manifold. Given that the intrinsic dimension of this latter sequence of manifolds begins to stabilize toward the end of denoising, we conjectured that there is a destabilizing factor that (i) primarily affects bottleneck representations and (ii) whose impact is stronger for larger time steps. One obvious culprit is the time step embedding vector which is added to input in the UNet at various points. 

To test this we fixed input to the UNet after denoising step 40 (out of 50), so that the only difference in denoising steps 41 through 50 is that the time step embedding vector is different. A plot of the intrinsic dimension of 500 bottleneck representations is shown in Figure \ref{fig-time-frozen}. We see that even when the input to the denoising process is frozen, intrinsic dimension continues to climb for the final 10 denoising steps. This strongly suggests that the time embedding vector is at least a significant contributor to the phenomenon described in Section \ref{sec:exper}. In Figure \ref{fig-time-frozen-comparison}, we give an example of the final output of the model when freezing is performed (right) and when it is not performed (left).

\begin{figure}[h]
\begin{center}
\includegraphics[width=0.49\linewidth]{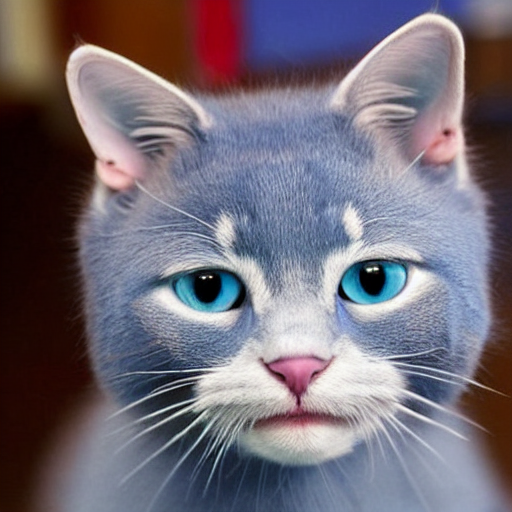}
\includegraphics[width=0.49\linewidth]{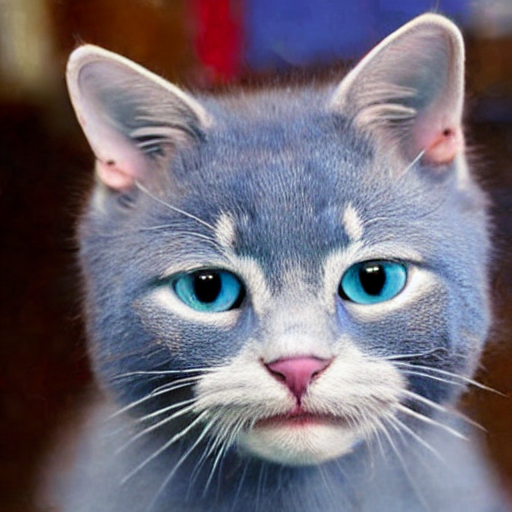}
\end{center}
\caption{The output of the model at the final (out of 50) denoising steps with the prompt `A cute blue cat' ({\textbf{Left}}) and the output of the model for 50 denoising steps when input is frozen at step 40 ({\textbf{Right}}).} \label{fig-time-frozen-comparison}
\end{figure}

\section{Intrinsic dimension of data and scaling laws}
\label{sec:envelope}

A number of works including but not limited to  \citep{pope2021the,sharmaNeuralScalingLaw2020,bahriExplainingNeuralScaling2021a}, have studied (both theoretically and empirically) the relationship between the intrinsic dimension of a dataset and the loss of a neural network trained on that dataset. Recently works trying to understand scaling have been extended to text-image models, both those trained with CLIP-style contrastive learning \citep{chertiReproducibleScalingLaws2022} and DALL-E-style autoregressive objectives \citep{aghajanyanScalingLawsGenerative2023}. Here, we briefly speculate on the relevance of these scaling laws to the experiments of Figure \ref{fig-perplexity}. 

As mentioned in \cref{sec:exper}, in the asymptotic regime of unlimited model parameters, \cite{bahriExplainingNeuralScaling2021a} relate the loss \(L \), dataset size \(D \) and data manifold intrinsic dimension as \(L = \mathcal{O}(D^{-1/d}) \). This would predict something like 
\begin{equation}
\label{eq:scaling-laws-lite}
    \log L \leq -\frac{1}{d}\log D  + \text{ constant}. 
\end{equation}
This inequality seems to suggest for \(D \) fixed, loss \(L \), and intrinsic dimension of the data manifold are monotonically related. Our result of Figure \ref{fig-perplexity} has a similar flavor, with the following notable differences:
\begin{itemize}
    \item Instead of the loss of the Stable Diffusion model in question, we have perplexity of an auxiliary language model.
    \item Instead of the intrinsic dimension of the Stable Diffusion training dataset, we have the intrinsic dimension of hidden features in certain layers. These are related to the intrinsic dimension of  output images, which the Stable Diffusion optimization process is trying to align with images in its training dataset, but nevertheless the two are different and one should be very cautious about conflating them.
    \item We are evaluating the Stable Diffusion model on a single prompt, rather than performing an average over a large dataset of prompts. It is not clear what plays the role of \(D \) in this analogy.
\end{itemize}
There is another way one might relate \cref{eq:scaling-laws-lite} to our experiments: if high perplexity prompts truly are more rare, we might imagine they effectively decrease the training data size \(D\) (since the model has seen them fewer times). This seems incompatible with the above analysis: if for some reason the loss stays constant, \(d \) would have to \emph{decrease} rather than increase as it does in Figure \ref{fig-perplexity}.

While quite simple, even simplistic, this discussion points to a number of avenues for future work: first, it is not obvious how to attempt something like the above back-of-the-envelope calculation using the multi-modal scaling laws of \citep{aghajanyanScalingLawsGenerative2023}. In a different direction, we were unable to find much on neural scaling laws for diffusion models (the closest we came across was \citep{caballeroBrokenNeuralScaling2023}).

\begin{figure}[h]
\begin{center}
\includegraphics[width=0.24\linewidth]{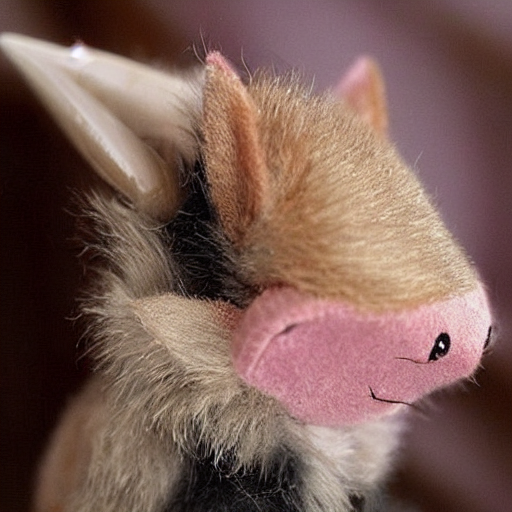}
\includegraphics[width=0.24\linewidth]{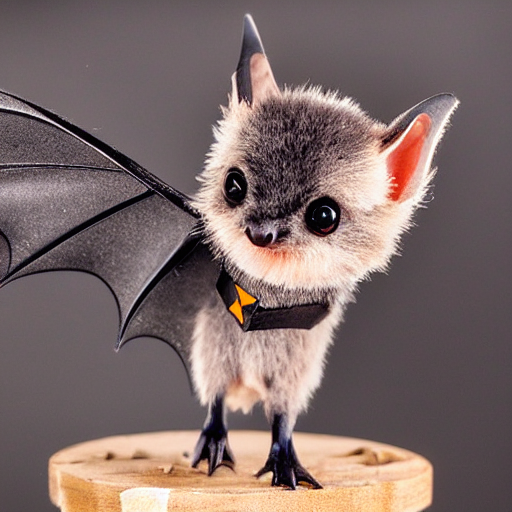}
\includegraphics[width=0.24\linewidth]{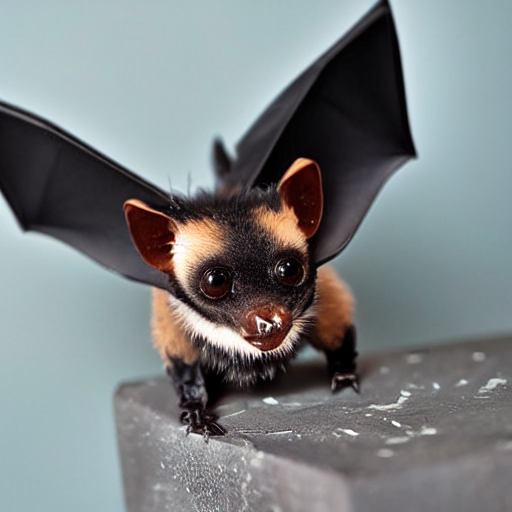}
\includegraphics[width=0.24\linewidth]{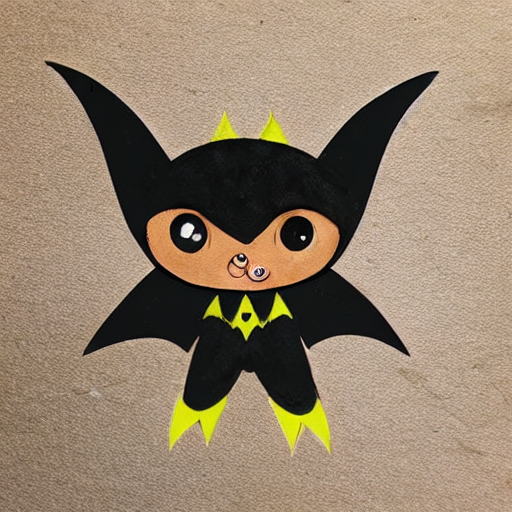}
\includegraphics[width=0.24\linewidth]{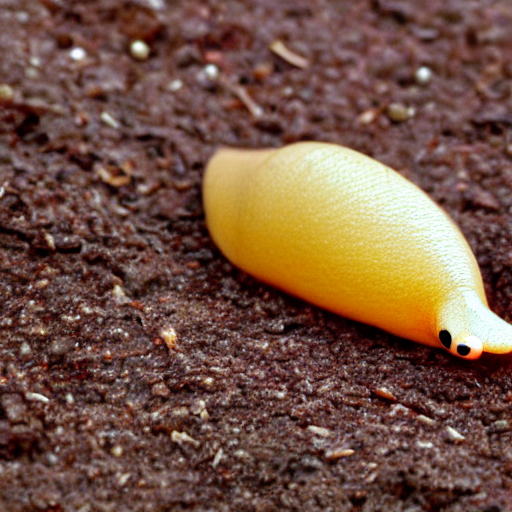}
\includegraphics[width=0.24\linewidth]{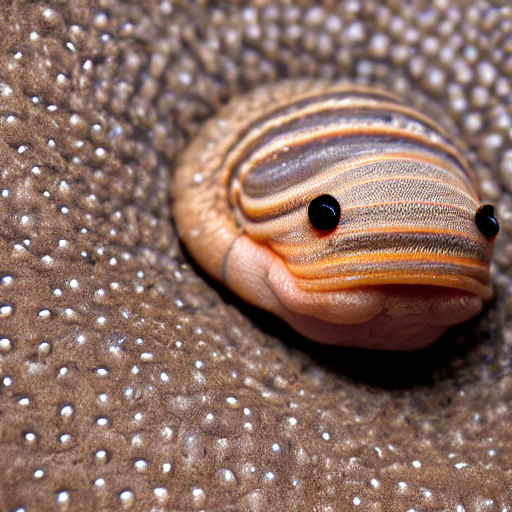}
\includegraphics[width=0.24\linewidth]{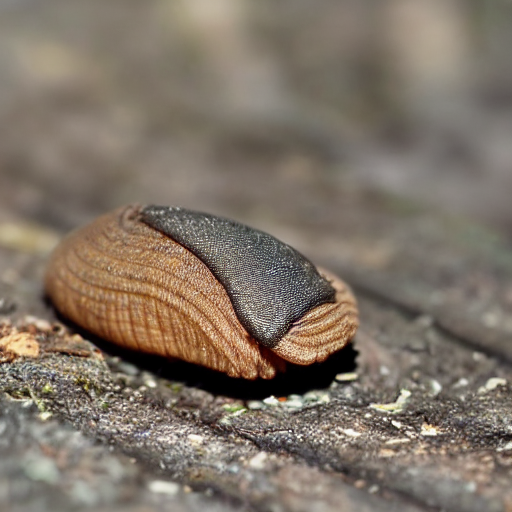}
\includegraphics[width=0.24\linewidth]{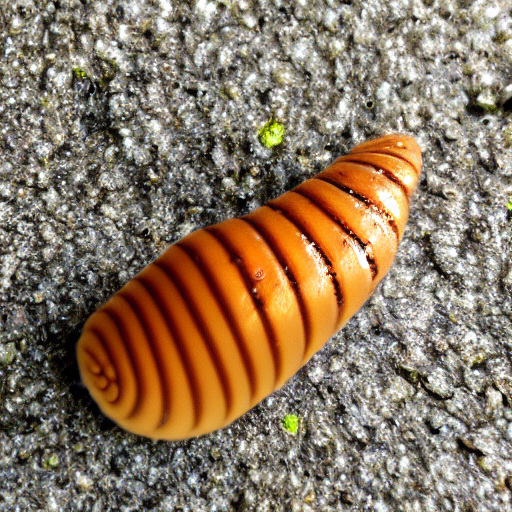}
\includegraphics[width=0.24\linewidth]{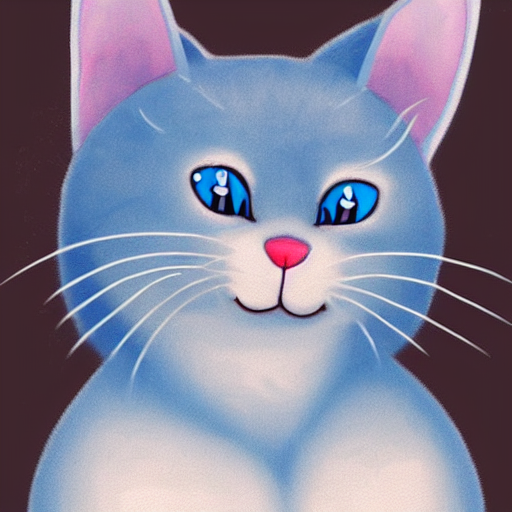}
\includegraphics[width=0.24\linewidth]{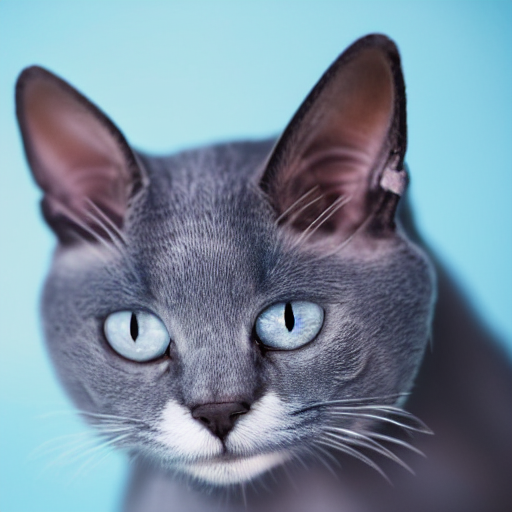}
\includegraphics[width=0.24\linewidth]{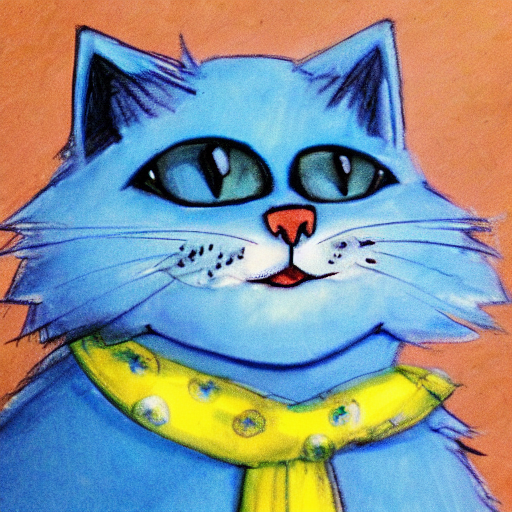}
\includegraphics[width=0.24\linewidth]{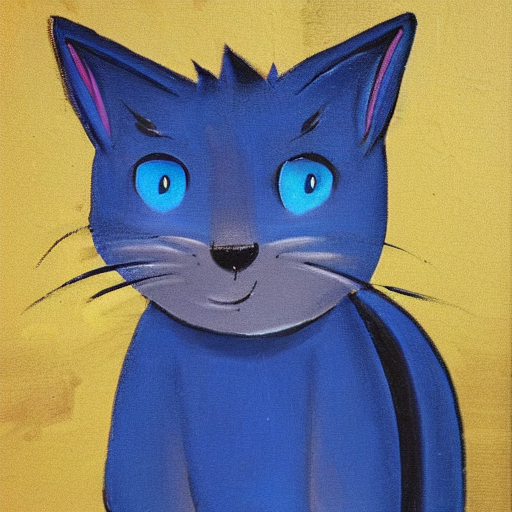}
\includegraphics[width=0.24\linewidth]{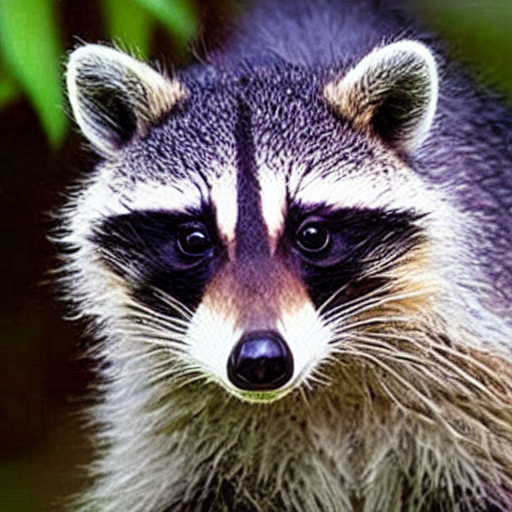}
\includegraphics[width=0.24\linewidth]{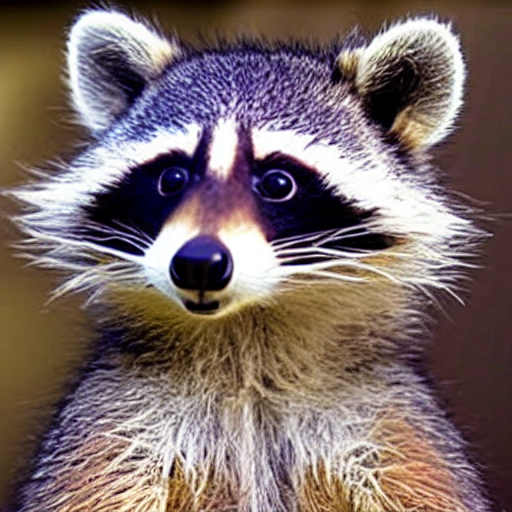}
\includegraphics[width=0.24\linewidth]{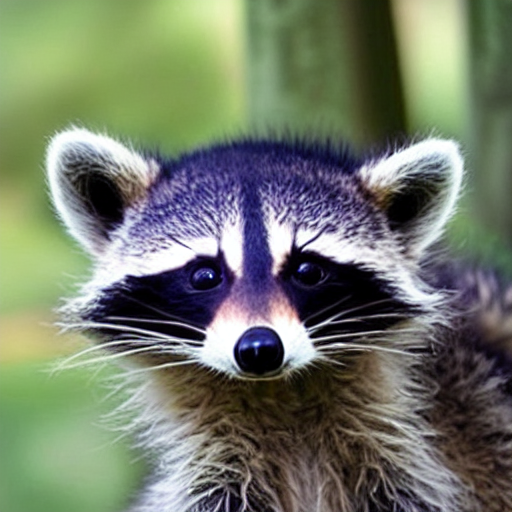}
\includegraphics[width=0.24\linewidth]{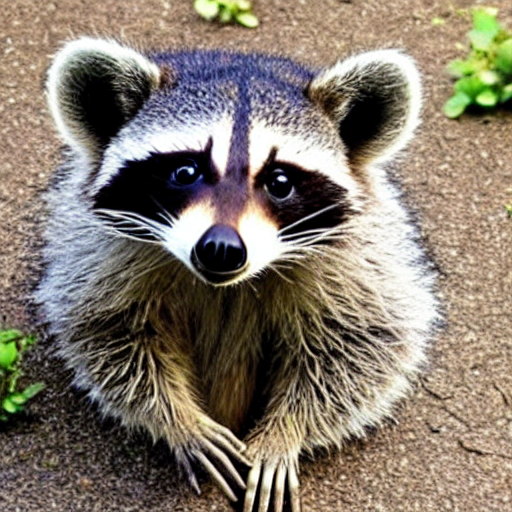}
\end{center}
\caption{Examples of images from the prompts: `A cute bat', `A cute slug', `A cute cat', and `A cute raccoon'.} \label{fig-time-frozen-comparison}
\end{figure}

\begin{figure}[h]
\begin{center}
\includegraphics[width=0.24\linewidth]{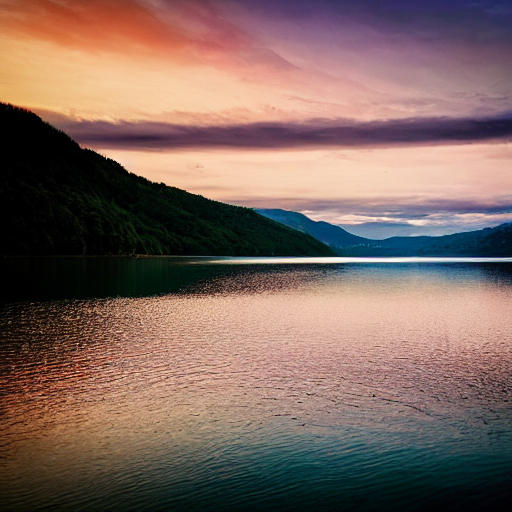}
\includegraphics[width=0.24\linewidth]{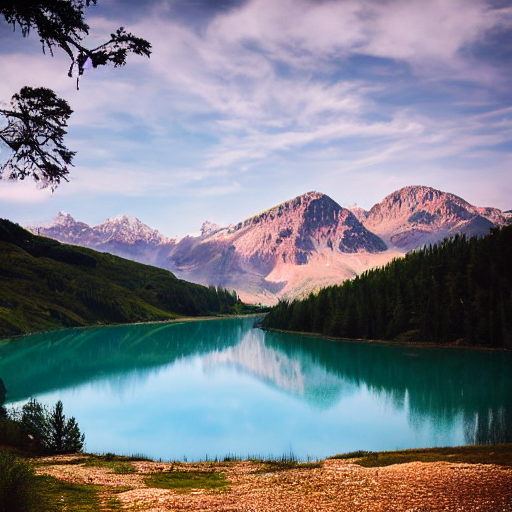}
\includegraphics[width=0.24\linewidth]{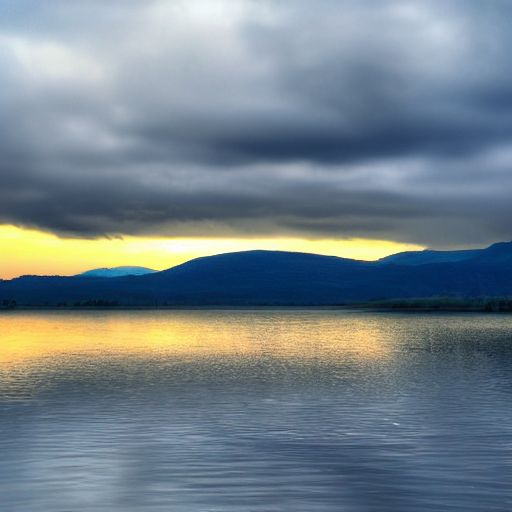}
\includegraphics[width=0.24\linewidth]{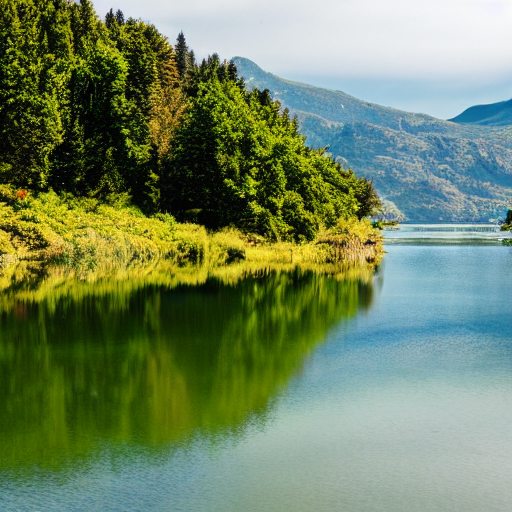}
\includegraphics[width=0.24\linewidth]{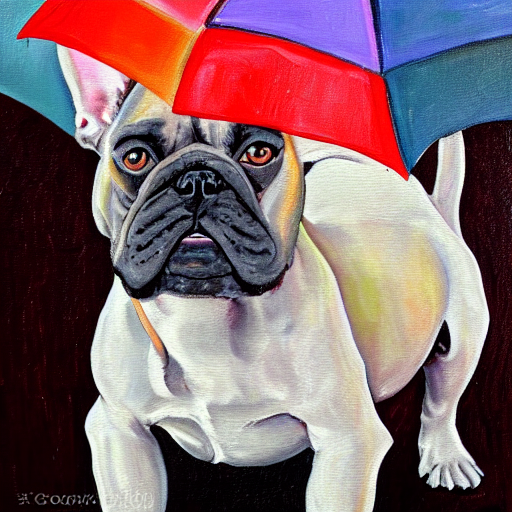}
\includegraphics[width=0.24\linewidth]{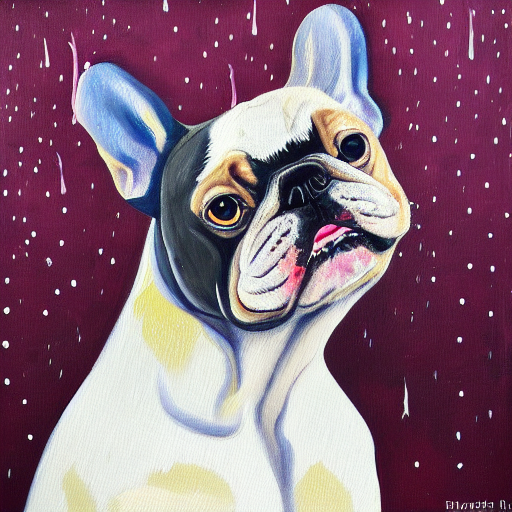}
\includegraphics[width=0.24\linewidth]{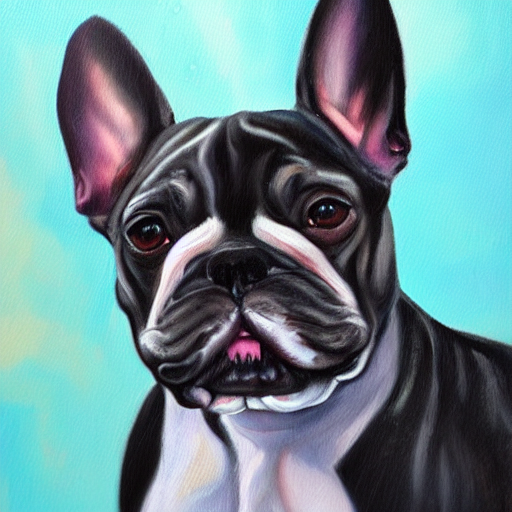}
\includegraphics[width=0.24\linewidth]{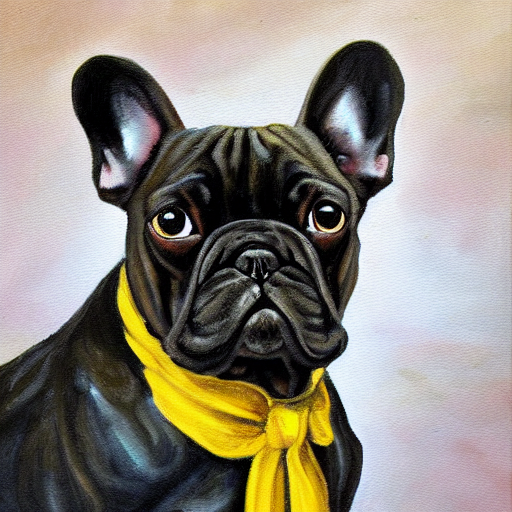}
\end{center}
\caption{Examples of images from the prompts: `A photograph of a beautiful lake' and `A painting of a French bulldog in the rain'.} \label{fig-time-frozen-comparison}
\end{figure}

\end{document}